# Using evolutionary machine learning to characterize and optimize co-pyrolysis of biomass feedstocks and polymeric wastes

Hossein Shahbeik[1,†], Alireza Shafizadeh[2,†], Mohammad Hossein Nadian[3], Dorsa Jeddi[2], Seyedali Mirjalili[4], Yadong Yang[5], Su Shiung Lam[6,1,*], Junting Pan[5,*], Meisam Tabatabaei[6,1,7,*], Mortaza Aghbashlo[2,1,*]

[1] *Henan Province Forest Resources Sustainable Development and High-value Utilization Engineering Research Center, School of Forestry, Henan Agricultural University, Zhengzhou 450002, China.*
[2] *Department of Mechanical Engineering of Agricultural Machinery, Faculty of Agricultural Engineering and Technology, College of Agriculture and Natural Resources, University of Tehran, Karaj, Iran.*
[3] *School of Cognitive Sciences, Institute for Research in Fundamental Sciences (IPM), Tehran, Iran*
[4] *Centre of Artificial Intelligence Research and Optimisation, Torrens University Australia, Brisbane, Australia.*
[5] *Institute of Agricultural Resources and Regional Planning, Chinese Academy of Agricultural Sciences, Beijing, 100081, PR China.*
[6] *Higher Institution Centre of Excellence (HICoE), Institute of Tropical Aquaculture and Fisheries (AKUATROP), Universiti Malaysia Terengganu, 21030 Kuala Nerus, Terengganu, Malaysia.*
[7] *Biofuel Research Team (BRTeam), Terengganu, Malaysia.*

* Correspondence:

Su Shiung Lam; Email: lam@umt.edu.my
Junting Pan; Email: panjunting@caas.cn
Meisam Tabatabaei: Email: meisam.tabatabaei@umt.edu.my
Mortaza Aghbashlo; Email: maghbashlo@ut.ac.ir

[†]These authors contributed equally.

**Abstract**

Co-pyrolysis of biomass feedstocks with polymeric wastes is a promising strategy for improving the quantity and quality parameters of the resulting liquid fuel. Numerous experimental measurements are typically conducted to find the optimal operating conditions. However, conducting co-pyrolysis experiments is highly challenging due to the need for costly and lengthy procedures. Machine learning (ML) provides capabilities to cope with such issues by leveraging existing data. Hence, the present study aims to introduce an evolutionary ML approach to quantify the (by)products of the biomass-polymer co-pyrolysis process. Multi-objective optimization is done to maximize pyrolysis oil production and minimize char/syngas formation simultaneously. A comprehensive dataset covering various biomass-polymer mixtures under a broad range of process conditions is compiled from the qualified literature. The database was subjected to statistical analysis and mechanistic discussion. The input features are constructed using an innovative approach to reflect the physics of the process. The constructed features are subjected to principal component analysis to reduce their dimensionality. The obtained scores are introduced into six ML models. Gaussian process regression model tuned by particle swarm optimization algorithm presents better prediction performance ($R^2 > 0.9$, MAE < 0.03, and RMSE < 0.06) than other developed models. The multi-objective particle swarm optimization algorithm successfully finds optimal independent parameters. Under optimal conditions, pyrolysis oil, char, and syngas yields are in the range of 70.9–75.3%, 7.23–21.5%, and 5.68–18.6%, respectively. The results demonstrate how ML can be employed to obviate the need for chemical-demanding, cost-intensive, and time-consuming co-pyrolysis experimental measurements.

| **Table of abbreviations:** | |
|---|---|
| ANFIS | Adaptive neuro-fuzzy inference system |
| C | Carbon |
| ELM | Extreme learning machines |
| GAM | Generalized additive model |
| GBR | Gradient boost regressor |
| GPR | Gaussian process regression |
| H | Hydrogen |
| H/C | Hydrogen-to-carbon |
| MAE | Mean absolute error |
| ML | Machine learning |
| MLPNN | Multi-layer perceptron neural network |
| MOPSO | Multiple objective particle swarm optimization |
| MSE | Mean square error |
| N | Nitrogen |
| N/C | Nitrogen-to-carbon |
| O | Oxygen |
| O/C | Oxygen-to-carbon |
| PCA | Principal component analysis |
| PSO | Particle swarm optimization |
| $R^2$ | Coefficient of determination |
| RMSE | Root-mean-square error |
| S | Sulfur |
| SVR | Support vector regressor |

## 1. Introduction

Biomass feedstocks can be converted to a wide spectrum of energy precursors, such as bio-oil, biocrude, biosyngas, bioethanol, biodiesel, biochar, and biomethane, through chemical, biochemical, and thermochemical technological platforms (Lim et al., 2022; Peter et al., 2023). Thermochemical conversion methods are preferred to chemical and biochemical approaches because of their flexibility concerning the feedstock and capability to deliver a wider range of value-added products. Various thermochemical conversion routes (i.e., direct combustion, hydrothermal processing, gasification, and pyrolysis) can convert biomass feedstocks into fuels and chemicals. The pyrolysis process is one of the promising thermochemical routes of taking profit from the energetic value of biomass (Hossain et al., 2017) due to its high process efficiency, substantial volume reduction ability, high energy recovery potential, zero-waste character, and capacity to produce a variety of value-added products (syngas, bio-char, and bio-oil) (Lu et al., 2018).

The endothermic pyrolysis process is carried out at temperatures in the range of 300–800 °C under vacuum, atmospheric, and pressurized conditions in the absence of oxygen or low oxygen concentrations (Ge et al., 2021). This process is classified into slow, fast, and flash pyrolysis based on the feedstock/product residence time and the reaction heating rate (Krishna et al., 2016). Among pyrolysis variants, fast pyrolysis has attracted a great deal of attention because of its capability to produce a liquid biofuel from biomass feedstocks within a short time (seconds to minutes) in one simple step (Xue and Bai, 2018). The pyrolysis oil has a higher energy density compared to the parent biomass. Nevertheless, the inherent oxygen content and hydrogen scarcity in the original biomass result in bio-oil with lower energy than petroleum-based fuels (Xue and Bai, 2018). In addition, bio-oil cannot be directly used as an alternative fuel due to its viscous, watery, acidic, corrosive, and reactive nature (Pinto et al., 2016). Accordingly, the chemically unstable bio-oil should be upgraded to transportation fuels and fine chemicals using hydrotreating, catalytic cracking, esterification, emulsification, and distillation techniques (Martínez et al., 2014; Shah et al., 2019). However, several downsides, such as process/technical complexities, high chemical/catalyst costs, and high investment/operating costs, make these upgrading methods technically unfeasible and economically unviable (Shah et al., 2019; Van Nguyen et al., 2019). A promising strategy to deal with the abovementioned issues is to co-pyrolyze biomass feedstocks with polymeric wastes.

Synthetic polymers are essential to today's human life due to their multifaceted properties, versatile characterizations, diverse applications, and affordable cost. During the past decades, the demand for synthetic polymers has constantly increased annually, proportional to population growth, rapid urbanization, economic development, and massive industrialization. Accordingly, a huge amount of short life cycle-polymeric waste is generated, leading to a primary concern in global waste treatment and a major environmental concern (Rotliwala and Parikh, 2011). Disposal of non-biodegradable polymeric waste in municipal solid waste landfills is occupying and depleting their space and causing serious air, water, and soil pollution (Izzatie et al., 2017; Jin et al., 2019). Pyrolysis is one of the most energetically efficient and environmentally friendly methods for dealing with recalcitrant polymeric waste. This process can completely convert and recover non-biodegradable organic compounds (Jin et al., 2019). The calorific value and hydrocarbon content of pyrolysis oil derived from polymeric waste are comparable with conventional gasoline (Stančin et al., 2021). Nevertheless, the thermal degradation of polymeric waste generates a significant amount of low-quality waxes and harmful compounds (i.e., polycyclic aromatic hydrocarbons, furans, dioxins, and benzenes) (Stančin et al., 2021; Tang et al., 2019). The generated waxes can cause various operational difficulties ranging from pipe clogging to reduced equipment lifespan (Tang et al., 2019). Co-pyrolyzing biomass feedstock with polymeric waste can facilitate the scission and breaking of the polymer chain, reducing the possibility of waxes and harmful compounds formation. In addition to improving pyrolysis oil quality, the co-pyrolysis approach is able to effectively address various environmental issues associated with polymeric waste disposal (Xue and Bai, 2018).

Factors such as the composition and nature of the biomass and synthetic polymer and the processing parameters (biomass blending ratio, temperature, and heating rate) can greatly affect the pyrolysis oil yield and its physicochemical properties (Rutkowski and Kubacki, 2006). Therefore, many experimental measurements must be scheduled to optimize the biomass blending ratio and reaction conditions in the co-pyrolysis of biomass with polymeric waste. Optimizing the co-pyrolysis process using the trial-and-error experimental routines is highly challenging due to their need for costly and lengthy procedures. On the other hand, the interaction mechanisms during the biomass-polymer co-pyrolysis process are not fully understood because of the occurrence of a series of consecutive and competing reactions (Chen et al., 2017). The interaction mechanisms vary feedstock by feedstock as well, making the process modeling more complex (Lu et al., 2018). In conclusion, it is quite challenging to characterize co-pyrolysis oil using theoretical methods, i.e., kinetic, chemical equilibrium, and computational fluid dynamic models (Yang et al., 2022). Therefore, an efficient data-driven method such as Machine Learning (ML) technology is vital to designing, optimizing, and scaling up biomass-polymer co-pyrolysis reactors. ML techniques can provide the chance to get the hidden nonlinear features and complex interactions of the biomass-polymer co-pyrolysis process by taking advantage of existing historical data. It also reduces human involvement in the process and inherently improves accuracy (khan et al., 2023; Suparmaniam et al., 2022). With the accumulation of experimental data on the biomass-polymer co-pyrolysis process, ML methods can effectively estimate the quantitative and qualitative characteristics of pyrolysis oil.

Recently, a few studies have used ML technology to model the kinetics of biomass-polymer co-pyrolysis. Table 1 tabulates some features of the ML approaches found in the published literature to model biomass-polymer co-pyrolysis. In general, the ML models reported in this research domain are case-specific, except for those developed by Wei et al. (2022)                    can only predict co-pyrolysis kinetics obtained using thermogravimetric analysis while disregarding the process products. Therefore, the present study aims to introduce a generic        approach to accurately predict the product distribution of the biomass-polymer co-pyrolysis process under various conditions. More specifically, six ML models are used to estimate the yield of pyrolysis oil, char, and syngas of biomass-polymer co-pyrolysis. The best-performing model is then used to optimize the process by maximizing pyrolysis oil yield and minimizing char and syngas yields. To the best of the authors' knowledge, this is the first attempt to use ML technology to characterize and optimize the co-pyrolysis process of biomass feedstocks and polymeric wastes.

Table 1. Features of the ML approaches found in the published literature to model biomass-polymer co-pyrolysis.

| Ref. | Feedstock types | ML model(s) | Model input(s) | Model output(s) | Model limitations |
|---|---|---|---|---|---|
| Yang et al. (2022) | Biomass pyrolysis coke, rapeseed cake | Artificial neural network | Blending percentage, operating temperature | Activation energy | - Developing the model based on limited biomass data points<br>-Considering only a limited number of effective input features<br>-Applicable only for predicting pyrolysis kinetics |
| Naqvi et al. (2019) | Rice husk, sewage sludge | Forward multi-layer perceptron | Blending percentage, operating temperature | Mass loss | - Applicable only to limited types of biomass feedstocks<br>-Considering only a limited number of effective input features<br>- Applicable only for predicting pyrolysis kinetics<br>- Limiting the model to a specific blending ratio<br>- Neglecting blending ratios of 0% and 100% |
| Ni et al. (2021) | Coal slime, coffee industry residue | Back propagation neural network | Temperature, heating rate, blending ratio | Mass loss | - Applicable only to limited biomass feedstocks<br>-Taking into account only a limited number of effective input features<br>- Applicable only for predicting pyrolysis kinetics<br>- Ignoring 0% and 100% blend ratios |
| Ni et al. (2022) | Coal gangue, coffee industry residue | Back propagation neural network | Temperature, heating rate, blending ratio | Mass loss | - Limited to certain types of biomass feedstocks<br>- Consideration of only a limited number of influential input features<br>-Applicable only for predicting pyrolysis kinetics<br>- Neglecting 0% and 100% blend ratios |
| Bi et al. (2021) | Sewage sludge, peanut shell | Artificial neural network | Temperature, heating rate, blending ratio | Remaining mass | - Limited to certain types of biomass feedstocks<br>-Taking into account only a limited number of effective input features<br>- Applicable only to pyrolysis kinetics prediction |
| Liew et al. (2021) | Corn cob, high-density polyethylene waste | Artificial neural network | Temperature, heating rate, blending ratio | Mass loss | - Ignoring 0 % and 100% blend ratios<br>- Limited to certain types of biomass feedstocks<br>-Taking into account only a limited number of effective input features<br>- Applicable only to pyrolysis kinetics<br>- Overlooking 0% and 100% blend ratios |
| Wei et al. (2022) | Coal, biomass | Random forest, extremely randomized trees | Ultimate analysis (CHNSO), volatile matter, ash content, fixed carbon content, temperature, heating rate, blending ratio | Mass loss | - Limited to certain types of biomass feedstocks<br>- Applicable only to pyrolysis kinetics prediction<br>- Ignoring blend ratios of 0% and 100% |

## 2. Research methodology

In this study, ML models were developed for the biomass-polymer co-pyrolysis process according to the flowchart shown in Figure 1. The papers published on co-pyrolysis of biomass feedstocks and polymeric wastes were first searched and selected. The eligible articles were screened based on their content and data. The reported data were extracted and evaluated. The valid datasets were introduced to six different supervised ML models, including multi-layer perceptron neural network (MLPNN), adaptive neuro-fuzzy inference system (ANFIS), extreme learning machines (ELM), support vector machine (SVR), generalized additive model (GAM), and Gaussian process regression (GPR). The best modeling option was selected based on several statistical criteria. The outcomes of the selected model (i.e., objective functions) were used in process optimization. Multiple objective particle swarm optimization (MOPSO) was applied to find optimum input parameters by maximizing pyrolysis oil yield while minimizing char and syngas yields.

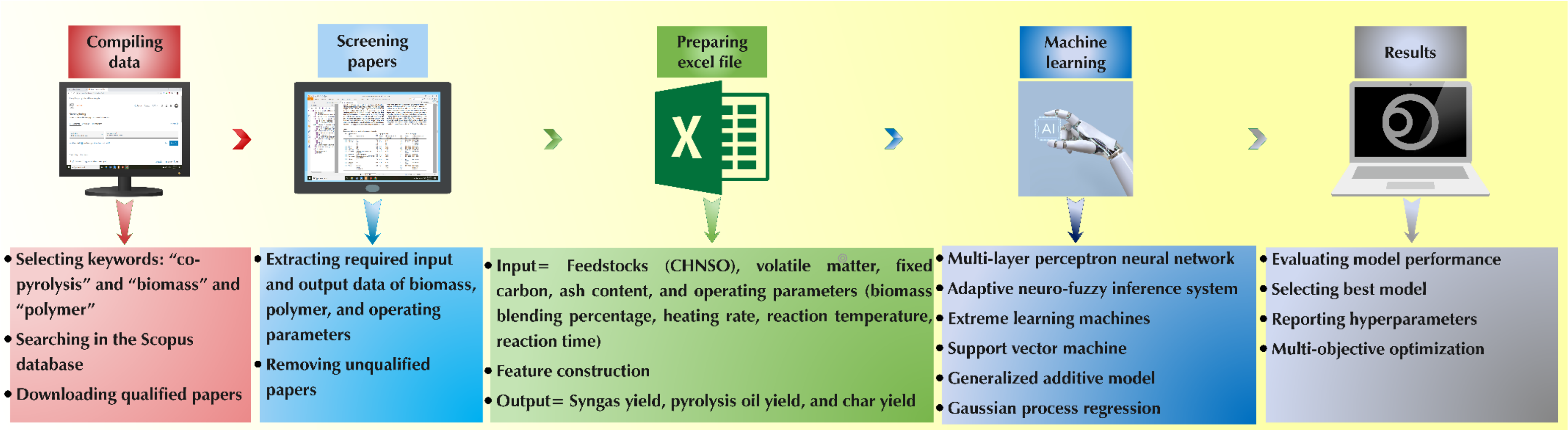

Figure 1. An overview of the research flowchart used in this study to develop ML models for biomass-polymer co-pyrolysis.

## 2.1. Data compilation

The Scopus database was searched and analyzed to obtain the data needed for developing ML models for the biomass-polymer co-pyrolysis process. The keywords of "co-pyrolysis" and "biomass" and "polymer" were first searched within the article title, abstract, and keywords in the Scopus database. The search engine initially introduced a total of 121 articles. Out of them, 49 suitable papers were chosen for more detailed screening. A set of 39 articles (34 biomass feedstocks and 22 polymeric wastes) had all the relevant parameters considered in this study. It was observed that ML models were developed to predict output parameters (pyrolysis oil, syngas, and char yields) based on elemental composition (CHNSO analysis) of both biomass feedstocks and polymeric wastes, proximate analysis (volatile matter, fixed carbon, and ash content) of both biomass feedstocks and polymeric wastes, and operating parameters (biomass blending percentage, heating rate, reaction temperature, and reaction time). The graphical data in the existing literature was extracted using image processing software (plot digitizer). The extracted data can be found in the "Raw data" sheet in "Supplementary Material".

Selecting suitable input parameters that can well reflect the physics of the process is a challenging task in ML modeling. This becomes even more challenging when modeling the co-pyrolysis process because biomass feedstocks and polymeric wastes are mixed with different natures. The biomass blending percentage alone cannot well reflect the effects of biomass and polymer composition on the output parameters. In other words, modeling the co-pyrolysis process based on 19 relevant parameters (see "Raw data" sheet in "Supplementary Material") might not lead to accurate models. More specifically, modeling results might be misleading at borderlines (biomass blending percentages of 0 and 100%). In addition, at biomass blending percentages between 0 and 100%, the modeling process is carried out based on a linear combination of biomass and polymer composition while neglecting their nature and essence. In order to deal with these issues, an innovative feature construction method was used in ML modelling (Eq. 1).

$$Z= [A\times X] + [(100\text{-}A)\times Y] + [(A\times X)^2 . ((100\text{-}A)\times Y)] + [(A\times X) . ((100\text{-}A)\times Y)^2] \quad (1)$$

where Z is the new input matrix with 32 columns developed based on biomass and polymer composition. A indicates the biomass blending percentage, and X and Y show the biomass and polymer composition matrix with 8 columns (CHNSO analysis, ash content, volatile matter, and fixed carbon), respectively. Each bracket creates eight separate columns. The $[A\times X]$ and $[(100\text{-}A)\times Y]$ matrixes are linear combinations of biomass and polymer composition. The $[(A\times X)^2 . (100\text{-}A)\times Y]$ and $[(A\times X) . ((100\text{-}A)\times Y)^2]$ matrixes are the nonlinear combinations of biomass and polymer composition. The 32 new columns plus three operating input parameters (35 columns) were used in the modeling process (see “Reconstructed input matrix” sheet in “Supplementary Excel file”). Accordingly, when the biomass blending percentage was 0%, the biomass composition was no longer considered in the modeling. In addition, when the biomass blending percentage was 100%, the polymer composition was not considered in the modeling. The nonlinear combinations could also reasonably consider the difference in nature and essence of biomass and polymer composition on outputs.

After data reconstruction, principal component analysis (PCA) was applied to reduce the input matrix size and feed the informative data to ML models. PCA is a multivariate statistical method employed in high-dimensional optimization problems to compress the dimensionality of the data source while holding a large fraction of the data characteristic by selecting the highest variance component in the data matrix (Heo et al., 2009). PCA has been chosen as it improves the accuracy of ML models with no substantially added computation cost. Upon applying PCA to the dataset, the reconstructed input dataset was decreased from 35 to 12 columns for all the output responses. A total of 99.5% of the insights present in the reconstructed data were accounted for by the PCA scores (see “PCA scores for pyrolysis oil”,

"PCA scores for char", and "PCA scores for syngas" sheets in "Supplementary Material"). Finally, the resulting PCA scores and dependent outputs were normalized between zero and one and used in ML modeling. For each output, an individual model was constructed. ML models were evaluated for their ability to predict biomass-polymer co-pyrolysis using the k-fold cross-validation approach. This method can examine and validate the model generalization ability unbiasedly. After the database was randomized, it was divided into k equivalent folds (Marcot and Hanea, 2021). The candidate ML model was trained using k-1 folds. The remaining fold was used to test the trained model. In order to ensure that all the collected data were used in training and testing, the process was replicated k times by choosing *k* sub-samples (Elmaz et al., 2020).

### 2.2. Modeling and optimization

The pyrolysis oil, syngas, and char yields of the biomass-polymer co-pyrolysis process were modeled using the MLPNN, ELM, ANFIS, GAM, SVR, and GPR approaches. MLPNN is a nonlinear mapping structure that mimics the human brain function. This model has been broadly used for modeling complex, noisy, and nonlinear data sets (Aghbashlo et al., 2012). MLPNN includes three layers (i.e., input, hidden, and output layers). The first and last layers are entirely linked to input parameters and desired outputs, respectively. The middle layer(s) are fully linked to the previous and subsequent ones. Each layer contains neurons that adjust the bias and weight values, thus optimizing the MLPNN structures. To adjust weights and biases, the MLPNN model uses several activation functions (i.e., ReLU, sigmoid, and tanh functions). The capability of the MLPNN model to solve stochastic and complex problems makes it one of the most widely used ML models (Darvishan et al., 2018; Shafizadeh et al., 2022).

ELM is a single-layer feed-forward MLPNN model. The method has been developed to eliminate the issues associated with gradient descent-based algorithms (i.e., long learning process). The ELM method can also make a more robust model than the MLPNN method. The algorithm can be used for classification and prediction purposes. It is also capable of self-adaptation that finds an optimal number of neurons in a neural network's hidden layer(s). This capability makes the ELM method superior to other ML techniques (Aghbashlo et al., 2016). ANFIS model is a hybrid ML model that benefits from the superior learning algorithms of the MLPNN model and the excellent estimation functions of the fuzzy inference systems (Aghbashlo et al., 2019). In addition, the weight values obtained from ANN can be explained in ANFIS, which is not possible in MLPNN (Karaboga and Kaya, 2019). The statistically flexible GAM model is based on consolidating the characteristic of generalized linear models and additive models (smooth functions of covariates). A shape function (i.e., boosted tree) is used in this model for each predictor or a pair of predictors. The GAM model can map nonlinear relationships between independent input parameters and dependent output responses. The GAM results are simply interpretable because the effects of individual shape functions on the desired outputs are separated.

The Support Vector Regression (SVR) is another ML used in this work that determines the optimal hyperplane separation between training samples. This hyperplane should maximize the margin between the training classes while decreasing the generalization error (Sabzekar and Hasheminejad, 2021). As a support vector machine`s sub-branch, the SVR is employed for regression analysis by applying respective linear or nonlinear kernel functions during modeling. The SVR is a promising technique for solving regression problems. This model is sensitive to noisy data or outliers because a small portion of data makes the training data hyperplane selection (Kavitha S et al., 2016). GPR applies the Gaussian process to training data before starting regression analysis. GPR is considered a non-parametric model that can effectively capture the nonlinearity in datasets. Unlike multiple linear regression, GPR does not require a fitting function declaration with an exact form due to its nature. In this method, the presented data are some points sampled from a multi-dimensional Gaussian distribution (Jiang et al., 2021). The GPR model can effectively analyze noisy data (Alodat and Shakhatreh, 2020). A brief overview of the mathematical background of the applied ML models can be found in the Supplementary Word File (Section S.1).

The hyperparameters of ML models were optimized and adjusted using the particle swarm optimization (PSO) algorithm. PSO is one of the widely used stochastic optimization algorithms. It is a black-box optimizer and does not require calculating the derivative of the problem. Furthermore, it avoids local optimal solutions that are common in single- and multi-objective real-world problems. These motivated our attempts to use it as the main optimization algorithm in this work. In this optimization method, a random population of solutions is first created. Each solution is considered to be a "particle flying" in an *n*-dimensional search space where *n* is the number of variables to be optimized. PSO updates the position of particles using a velocity vector. When calculating the velocity, each particle search for the optimum position according to its best previous position and the optimal position found by all particle. This iterative process of calculating velocity, updating position, and re-evaluating particles using the objective function is iteratively continued until the satisfaction of an end condition, which can be reaching a certain accuracy level or hitting a pre-determined maximum number of iterations (Shafizadeh et al., 2022). The PSO algorithm is able to solve problems with a single-objective nature. To solve problems with more than one objective, MOPSO has been developed in the literature. In MOPSO, particles are compared using the concept of Pareto optimality, and a set of solutions representing the best trade-offs between the objectives are found (a.k.a estimated Pareto optimal solution set). The MOPSO algorithm was also used to optimize the biomass-polymer co-pyrolysis process simultaneously. In the Supplementary Word File (Tables S2 and S3), pseudo-codes for PSO and MOPSO algorithms are provided. The characteristics of the PSO and MOPSO algorithms used herein were similar to those in the previous study (Atarod et al., 2021).

## 3. Results and discussion

### 3.1. Data analysis

The statistical analysis was conducted to comprehend the details and characteristics of biomass-polymer co-pyrolysis variables. Figure 2 depicts the statistical analysis of the collected data. In general, the carbon content of biomass feedstocks (34.1–79.8%) was significantly lower than those of waste polymers (38.3–92.4%). The higher organic matter content of waste polymers resulted in the production of carbon-rich alternative fuel during the co-pyrolysis process (Figure 2A). Moreover, the volatile matter of waste polymers (60.7–100%) was higher than biomass feedstocks (9.60–97%), resulting in a positive synergistic effect on improving pyrolysis oil quality during the co-pyrolysis process (Ansari et al., 2021). The higher ash level of biomass feedstocks (with a median value of 5.07%) was mainly rooted in their cultivation environment. The higher volatile matter and the lower ash content of waste polymers rendered them ideal feedstocks to enhance the quality of the biomass-derived pyrolysis oil in the co-pyrolysis process. The elemental composition of different waste polymers indicated that they were mainly composed of carbon and hydrogen with low oxygen, nitrogen, and sulfur. Accordingly, waste polymers were good sources of liquid hydrocarbons, and their co-pyrolysis with biomass feedstocks could quantitatively and qualitatively improve the resultant liquid fuel because of the synergistic interactions between biomass and polymeric waste (Uzoejinwa et al., 2018).

The reaction temperature ranged from 350 to 1100 °C, corresponding to a wide range of possible operating temperatures (Figure 2B). The median heating rate and reaction time stood at 15 °C/min and 30 min, respectively. In terms of the output targets, the highest contribution to resultant products belonged to the liquid fuel yield with a median value of 45.6%, followed by char yield (22.1%) and syngas yield (25.9%) (Figure 2C). The range of product distribution was very large owing to the wide range of feedstock compositions and operating conditions. It should be noted that the ML model developed based on such an inclusive database could effectively generalize and optimize the co-pyrolysis process of biomass and polymeric waste.

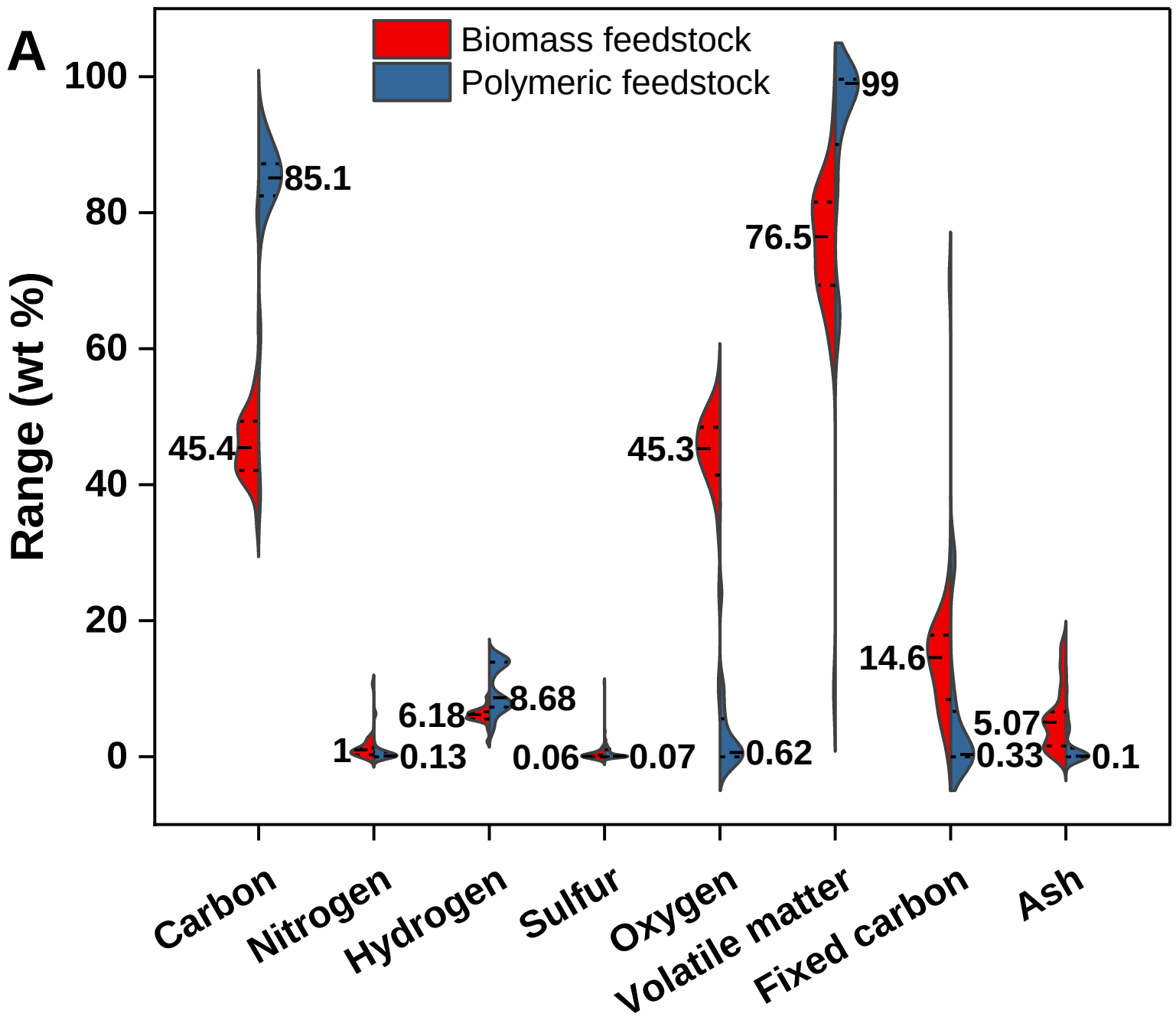

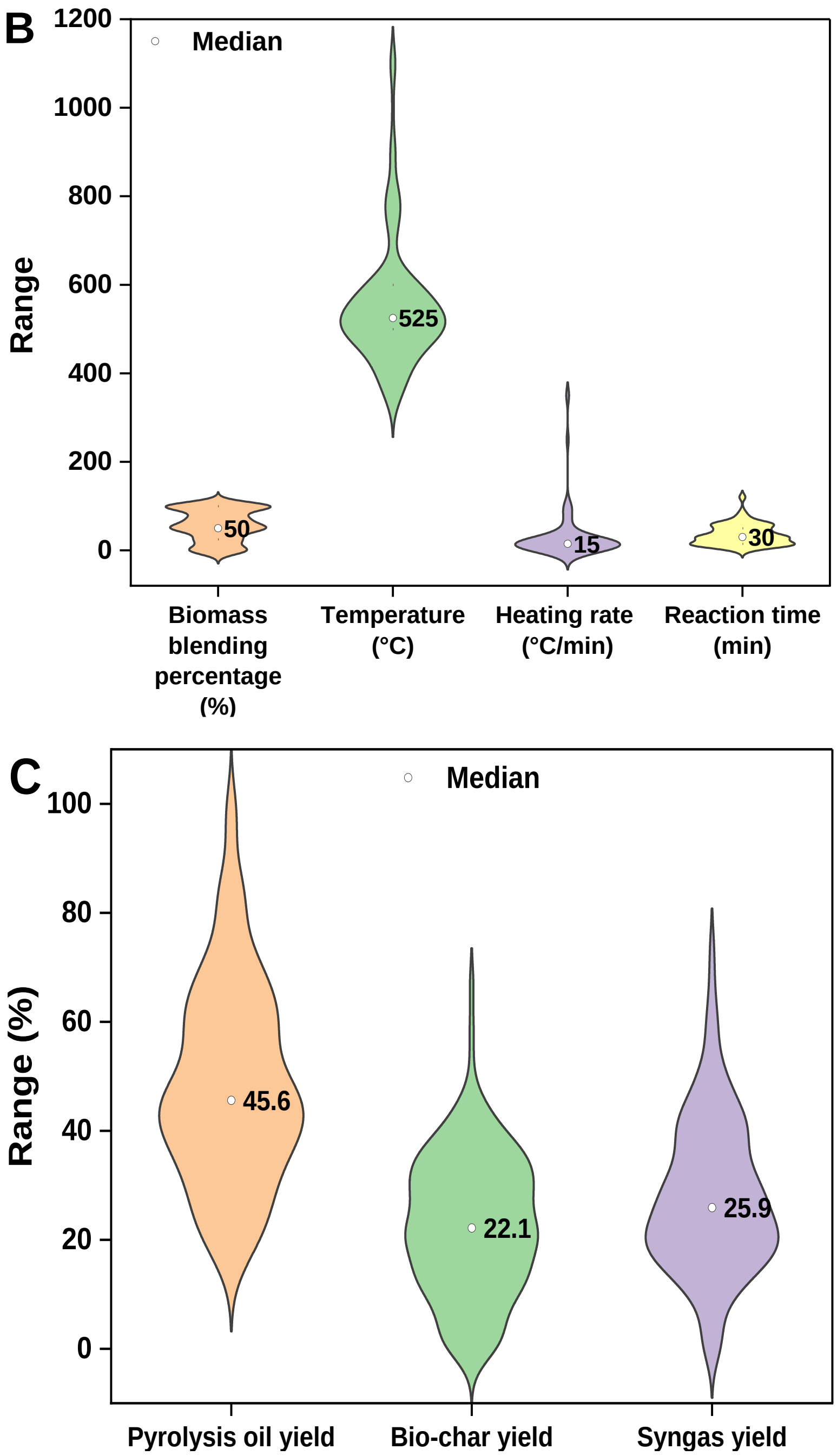

Figure 2. Violin plot analysis of collected data. (A) feedstock composition, (B) reaction conditions, and (C) pyrolysis oil/syngas/char yield.

A preliminary relevance degree among the input features and output targets in the co-pyrolysis process was investigated based on the Spearman correlation analysis. This analysis is a non-parametric correlation type based on the ranked values rather than the raw data, measuring the magnitude and orientation of the relationship between two parameters according to a monotonic criterion (George et al., 2021). The values lie between -1 (complete negative correlation) and 1 (complete positive correlation), indicating a relatively low or high impact. Figure 3 depicts the correlation matrix among the inputs and outputs of the biomass-polymer co-pyrolysis process. The biomass blending percentage (biomass quantity to the whole feedstock) was negatively correlated with pyrolysis oil yield. Generally, decreasing the biomass blending percentage (or increasing the quantity of polymeric waste in the feedstock) increased the liquid product yield due to the higher amount of volatile matter in polymeric compounds while lowering the quantities of gaseous and solid products.

The reaction temperature was negatively associated with pyrolysis oil yield. Increasing the reaction temperatures (particularly with rapid heating rates) favored the evolution of gaseous compounds by promoting the primary decomposition of feedstock, the secondary cracking of volatiles, and the decomposition of char residues into incondensable gases (Chintala, 2018). This result was also attributed to the advancement of cracking reactions that converted longer and heavier molecules into smaller molecules, enhanced the volatilization of waxes, and depolymerized the evolved products (Wan Mahari et al., 2021). However, it should be highlighted that elevating the reaction temperature to a certain value (550–750 °C) increased the pyrolysis oil yield by promoting the thermal cracking of feedstock compounds. The heating rate was also negatively correlated with pyrolysis oil yield. Increasing the process heating rate greatly enhanced the heat and mass transfer, thus promoting the bond-scission reactions and boosting the tar and gas yields. The higher heating rate also lowered the occurrence of repolymerization and condensation reactions by shortening the residence time of primary volatiles (Foong et al., 2021). In addition, prolonged reaction time produced more waxes and char residues because of the carbonization of the materials into solid products, typically occurring at lower process temperatures. The prolonged reaction time at higher process temperatures promoted the secondary pyrolysis reactions to generate more gases while decreasing the liquid oil yield (Foong et al., 2021).

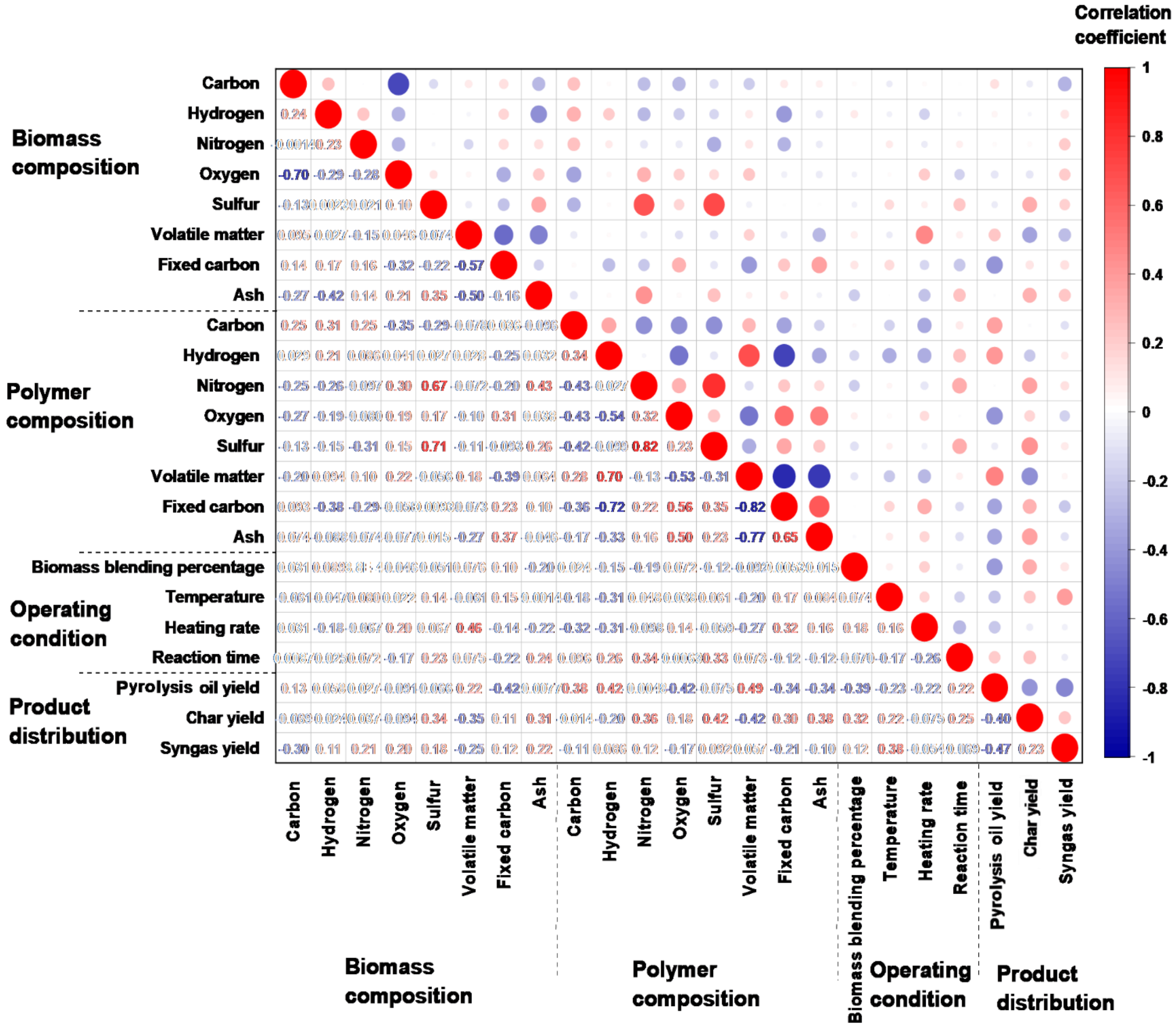

Figure 3. Correlation matrix between inputs and outputs.

Figure 4 illustrates the van Krevelen diagrams of different biomass and polymeric waste feedstocks. The average O/C and N/C ratio values of biomass feedstocks and polymeric wastes were 0.736 & 0.026 and 0.04 & 0.009, respectively. Polymeric compounds had lower O/C and N/C ratios than biomass feedstocks. Accordingly, polymeric compounds could effectively neutralize the negative effects of the inherent higher oxygen and nitrogen contents of biomass feedstocks in the co-pyrolysis process. This unique feature of waste polymers increased the homogeneity and consistency of the resultant products while minimizing coke formation during the co-pyrolysis process (Wang et al., 2021). The higher effective H/C ratio ($H/C_{eff}$) of polymeric wastes, particularly high/low-density polyethylene and polypropylene, could effectively enhance the quality and calorific value of the liquid fuel produced through the co-pyrolysis process (as in Figure 5). The increased calorific value was attributed to the lower water content and oxygen concentration of the resultant liquid fuel. The polymeric feedstock with a high $H/C_{eff}$ ratio (because of the higher amount of hydrocarbons containing linear and aromatic compounds in their composition) could function as a hydrogen source in the co-pyrolysis process, significantly increasing the hydrocarbon content of liquid products (Hassan et al., 2016). It is noteworthy that the polymeric wastes could promote the hydrodeoxygenation of biomass-derived oxygenated compounds during the co-pyrolysis process through hydrogen donation reaction (e.g., producing hydrogen radicals), thus enhancing the formation of hydrocarbons (Abomohra et al., 2021). More specifically, the increment in hydrocarbons was due to the Diels–Alder reaction (a cycloaddition reaction between conjugated dienes and alkenes to produce cyclohexene derivatives) between biomass-derived furans and polymer-derived olefins (Hassan et al., 2016). Furthermore, polymeric wastes could improve hydrocarbon selectivity while minimizing catalyst deactivation in the co-pyrolysis process, thanks to their higher $H/C_{eff}$ ratio (Ansari et al., 2021).

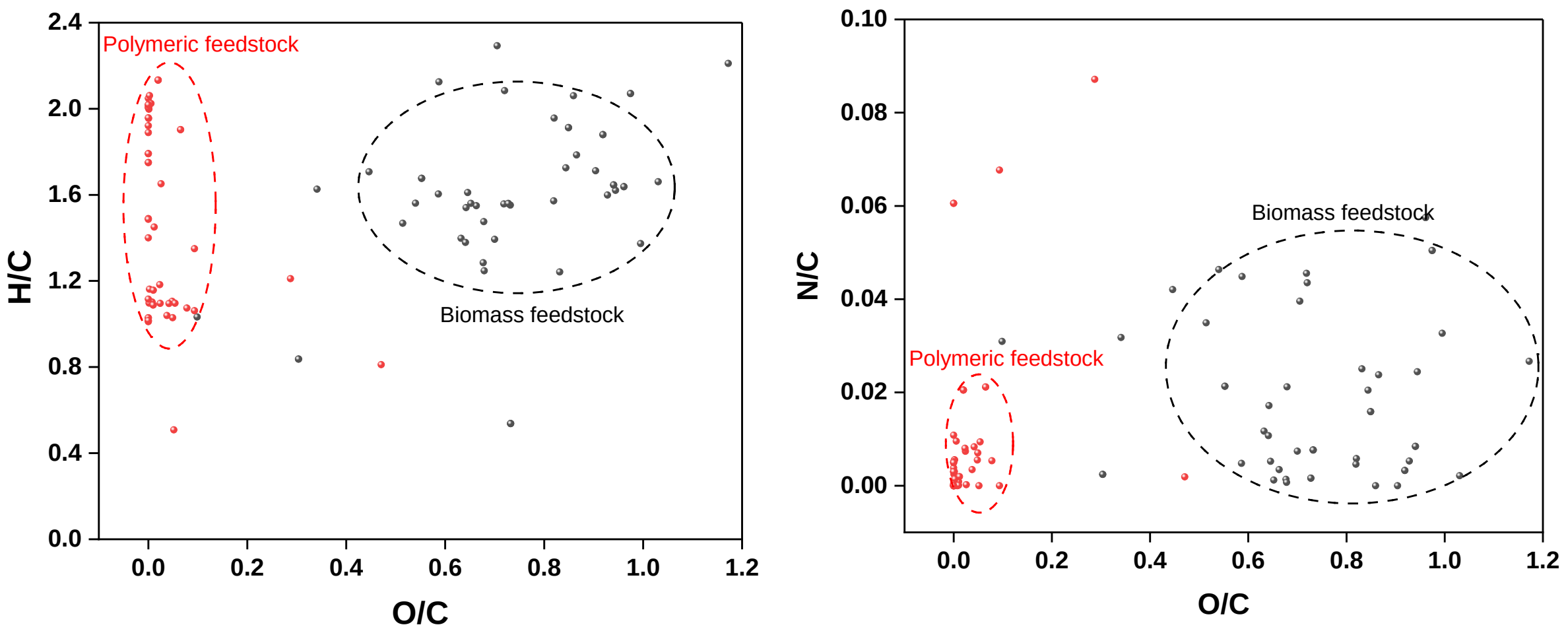

Figure 4. Elemental ratios of biomass feedstocks and polymeric wastes.

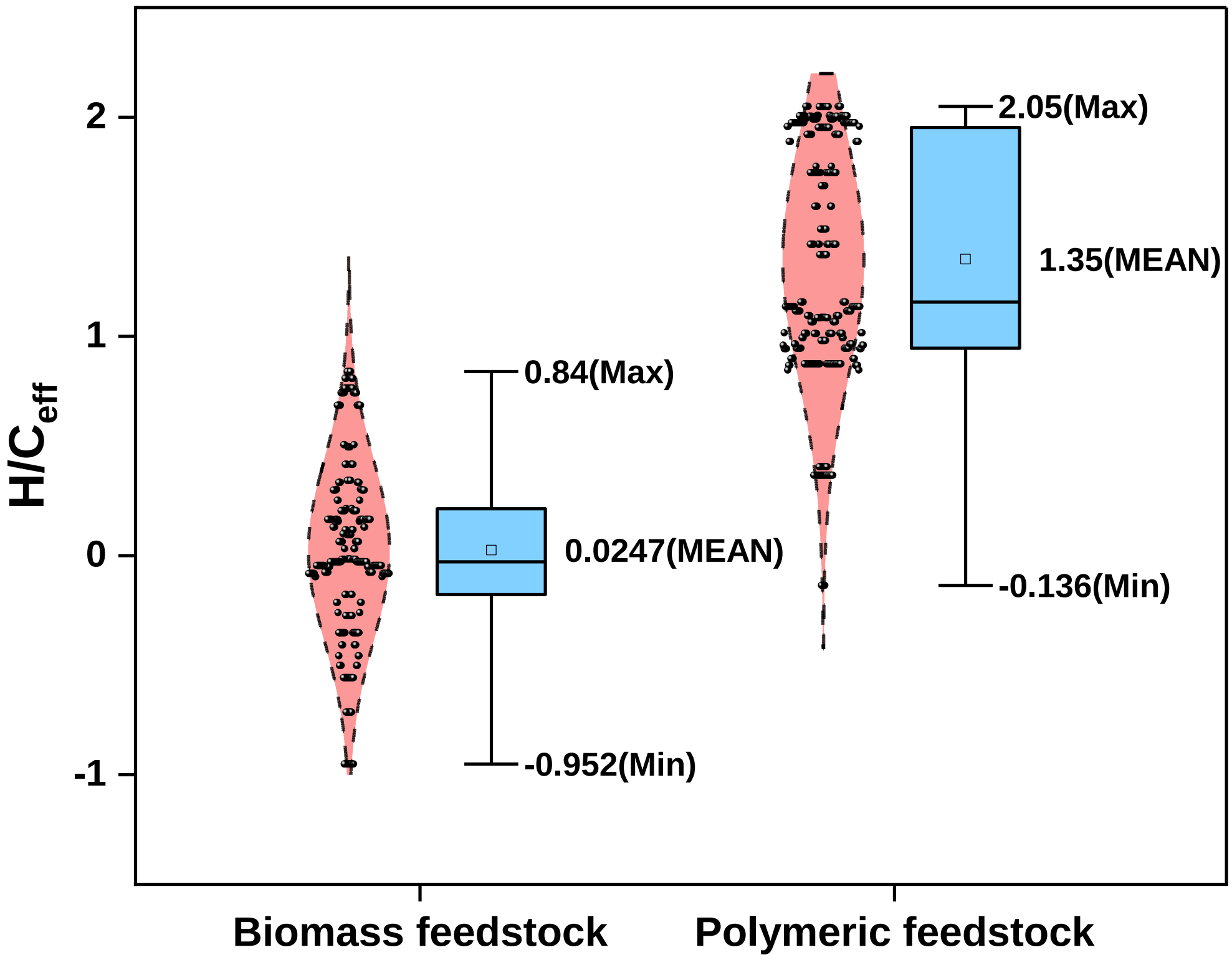

Figure 5. $H/C_{eff}$ ratio of biomass feedstocks and polymeric wastes. The $H/C_{eff}$ ratio is calculated as follows: $\frac{H}{C_{eff}} = \frac{H-2O-3N-2S}{C}$. H, O, N, S, and C are the mole percentages of hydrogen, oxygen, nitrogen, sulfur, and carbon, respectively. The $H/C_{eff}$ ratio is based on the assumption that all the heteroatoms are completely converted to $H_2O$, $NH_3$, and $H_2S$.

The PCA method was used once again to reduce the dimension of the feature space and identify the most relevant features of the dataset. The first PCA component amounted to 21.40% of the total variance. This value was 14.86% for the second PCA component (Figure 6A). Nearly 60% of the total variance derived from the dataset was attributable to the first four PCA components. The first PCA component comprised attributes such as nitrogen, oxygen, fixed carbon, and ash contents of polymeric feedstock and reaction temperature (Figure 6B). The second PCA component consisted of carbon content and fixed carbon of biomass feedstock. Both of these factors demonstrated how important feedstock composition was to biomass-polymer co-pyrolysis. Figure 6C shows the correlation of input/output features with the first and second PCA components. A meaningful relationship can be evidently observed between the input parameters and the quantity of the resultant liquid fuel, char, and syngas. More specifically, the volatile matter of polymeric wastes could positively affect the liquid fuel yield since, during the co-pyrolysis process, it migrates into the liquid fraction. Increasing the reaction temperature could accelerate the secondary decomposition and cracking reactions, producing more O- and N-containing gaseous compounds and increasing syngas yield. The char yield was increased by increasing the ash content of both biomass and polymer at the expense of reduced liquid fuel yield because of the fewer organic compounds and higher inorganic minerals in ash-rich feedstocks. It should be noted that primary and secondary reaction pathways yielded char during the pyrolysis process. The primary char was produced *via* dehydration, while recombining reactive and volatile fragments through cross-linking and condensation results in the secondary one (Gouws et al., 2021). Notably, these relationships might vary slightly due to variations in feedstock properties and reaction parameters.

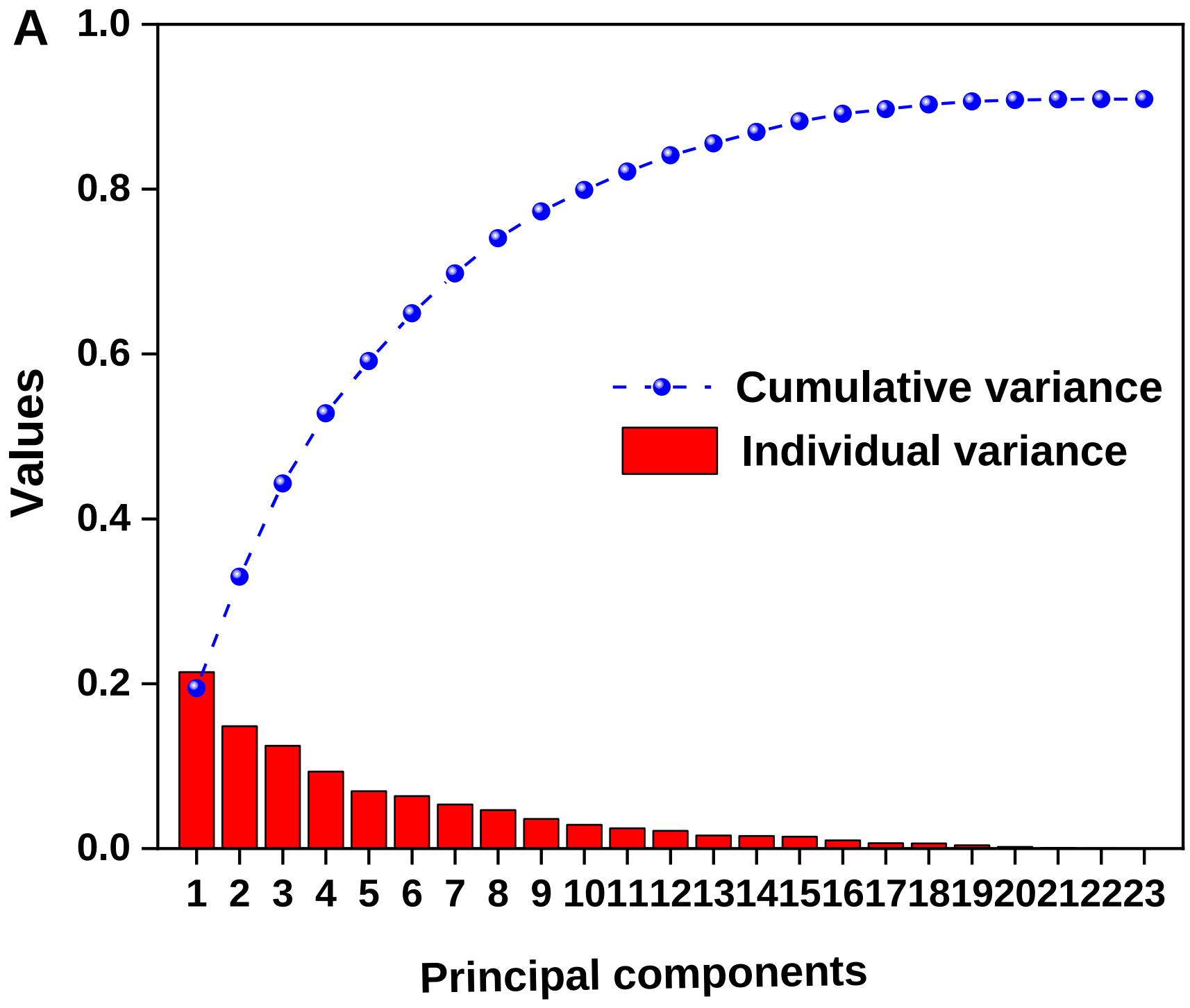

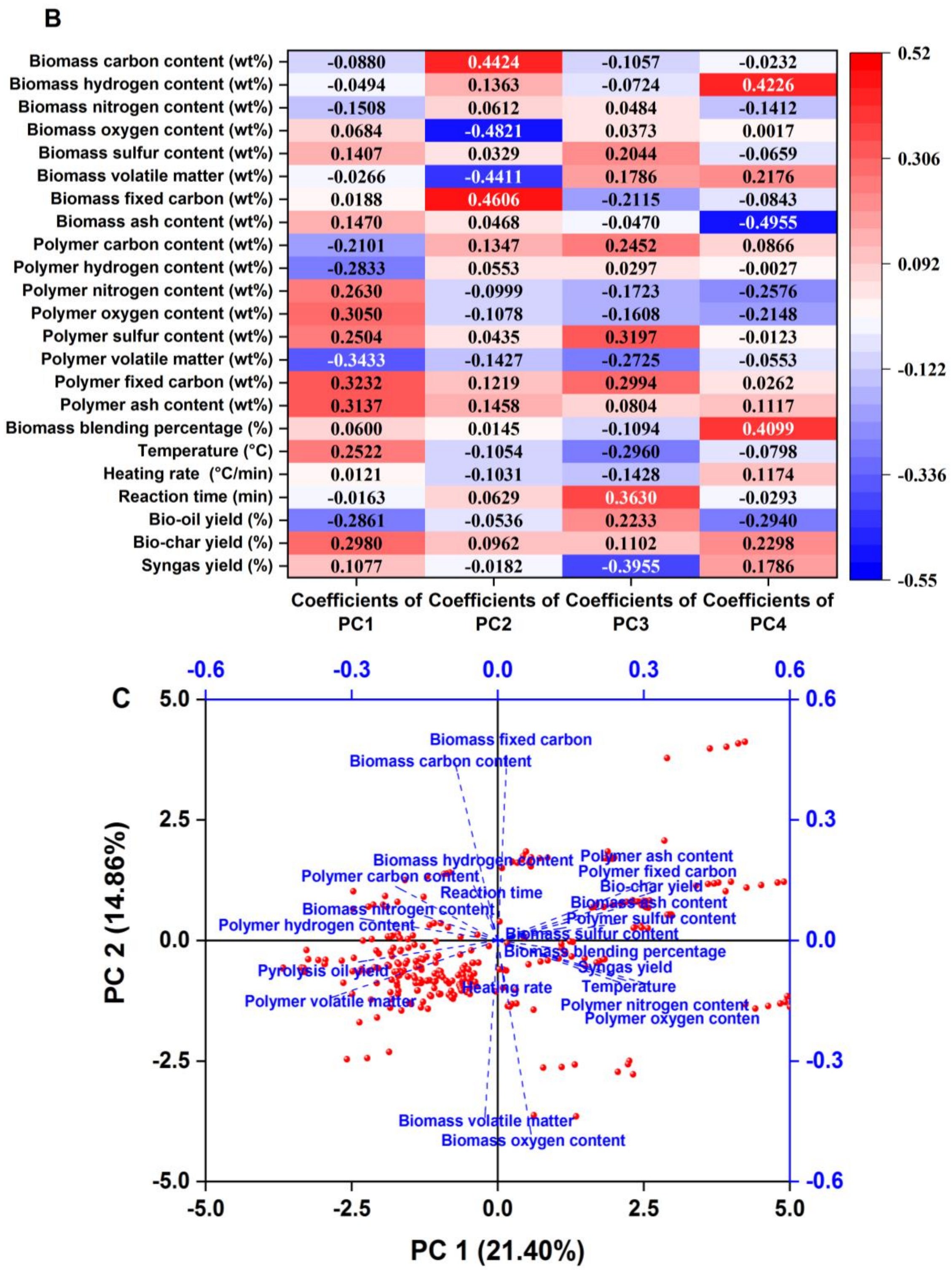

Figure 6. PCA analysis of the collected data. (A) Variance of each PCA component, (B) relationship between input/output features and the top-four PCA components, and (C) effect of inputs on pyrolysis oil, syngas, and char yields.

The relationship between the most important variables of the co-pyrolysis process (such as temperature, biomass blending percentage, and reaction time) and the product distribution (pyrolysis oil, syngas, and char yields) are depicted using the contour plot in Figure 7. The pyrolysis temperature, biomass blending percentage, and reaction time ranging from 550 to 750 °C, from 2 to 10%, and from 30 to 60 min, respectively, could provide more pyrolysis oil. Generally speaking, the pyrolysis temperature must be raised to an optimal value in the co-pyrolysis process by considering the amount of the released volatiles and the secondary cracking reactions. Under such an elevated temperature, the formation of intermediate byproducts (e.g., radicals, carbocations, and hydrogen donors) must be optimized to maximize liquid fuel yield without generating too many non-condensable gases (Gouws et al., 2021). However, elevating the pyrolysis temperature to above the optimum value could speed up the primary decomposition of feedstock, the secondary decomposition of volatiles, and the decomposition of char residues. These could, in turn, result in the release of more gaseous products while negatively lowering pyrolysis oil yield (Lu et al., 2009). The longer reaction time could also promote repolymerization and recondensation reactions and decrease pyrolysis oil yield by increasing char yield (Guedes et al., 2018).

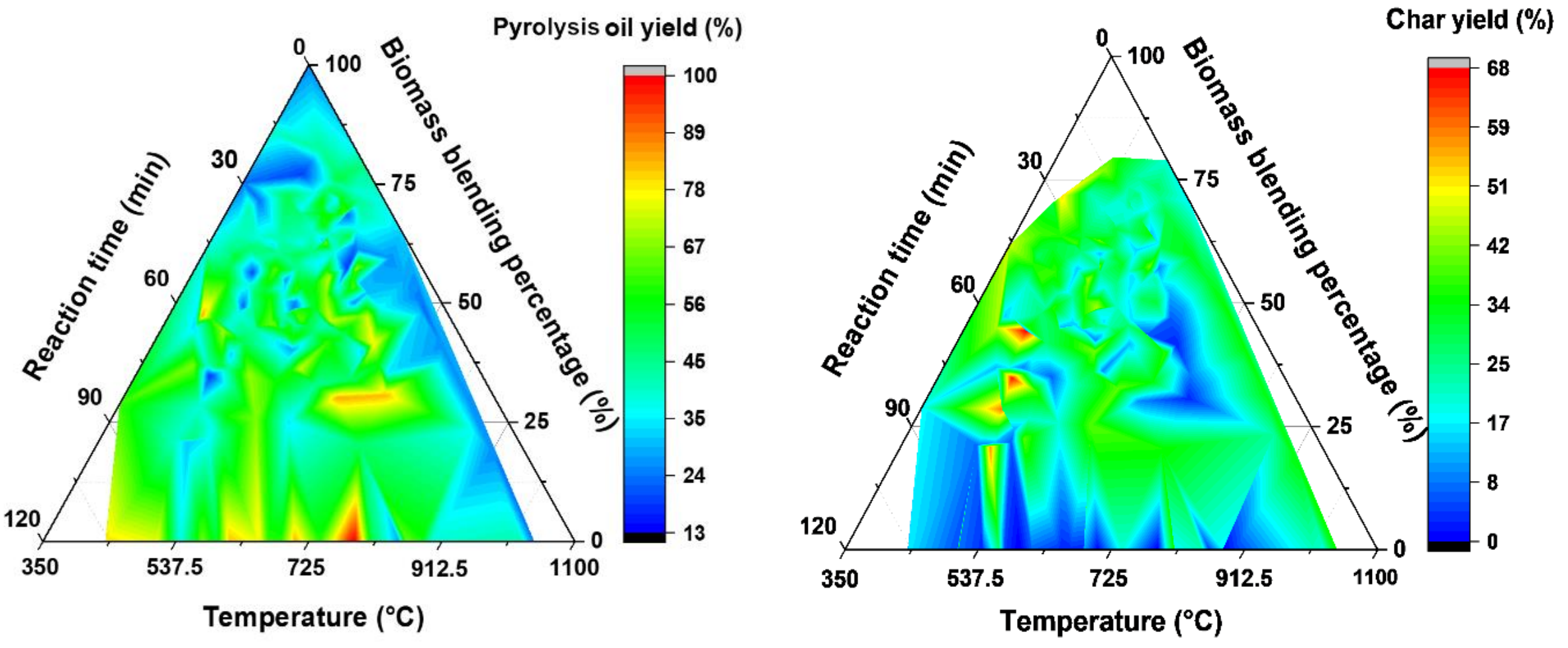

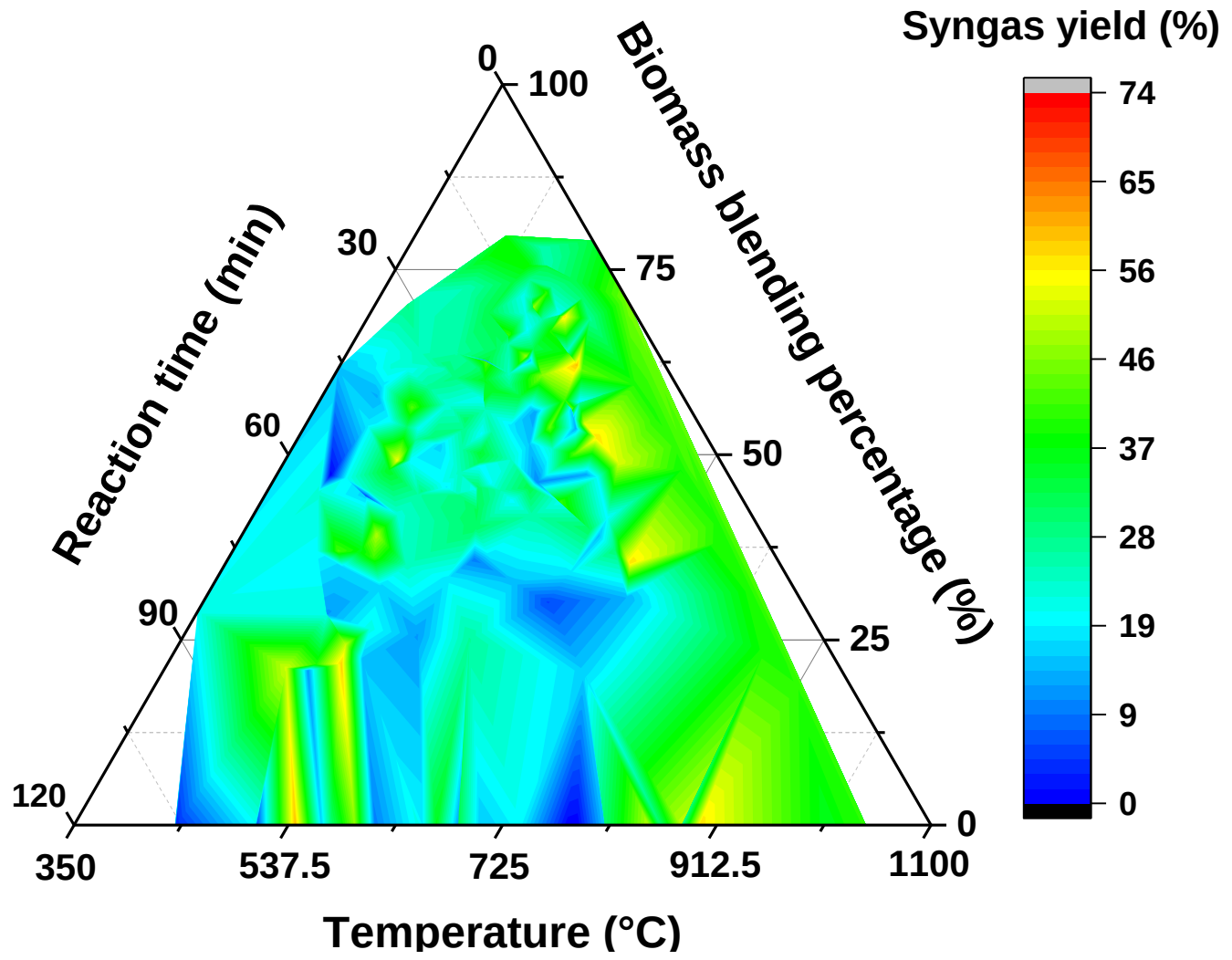

Figure 7. Relationship between the most important variables of the co-pyrolysis process (such as temperature, biomass blending percentage, and reaction time) and the product distribution (pyrolysis oil, syngas, char yields) using the contour plot.

### 3.2. Biomass-polymer co-pyrolysis mechanism

In biomass-polymer co-pyrolysis, free radical interactions are considered the most common mechanism (Figure 8). This mechanism includes radical initiation, secondary radical formation (depolymerization, hydrogen transfer, monomer formation, and isomerization), and termination by recombination or disproportionation of radicals (Engamba Esso et al., 2022). In co-pyrolysis, carbon-rich polymeric waste can act as a hydrogen donor to oxygen-rich biomass, thereby balancing the oxygen, carbon, and hydrogen contents of the feedstock (Brebu et al., 2010; Hassan et al., 2020). In fact, by transferring hydrogen from polymer chains to biomass-derived radicals, the primary products of cellulose decomposition can be stabilized, enhancing pyrolysis oil yield while lowering char yield (Lu et al., 2018). On the other hand, the thermally unstable radicals derived from biomass can promote the degradation of synthetic macromolecules (Brebu et al., 2010). The higher mineral content (i.e., potassium) of biomass can also catalyze the decomposition of both biomass and synthetic macromolecules in the co-pyrolysis process (Jin et al., 2019). Generally, radical interactions between polymer and biomass can enhance the quality and quantity of pyrolysis oil (Hassan et al., 2020). The resultant high-quality pyrolysis oil can subsequently be upgraded into motor fuels and valuable chemicals with lower costs and environmental impacts than biomass-derived bio-oil (Van Nguyen et al., 2019).

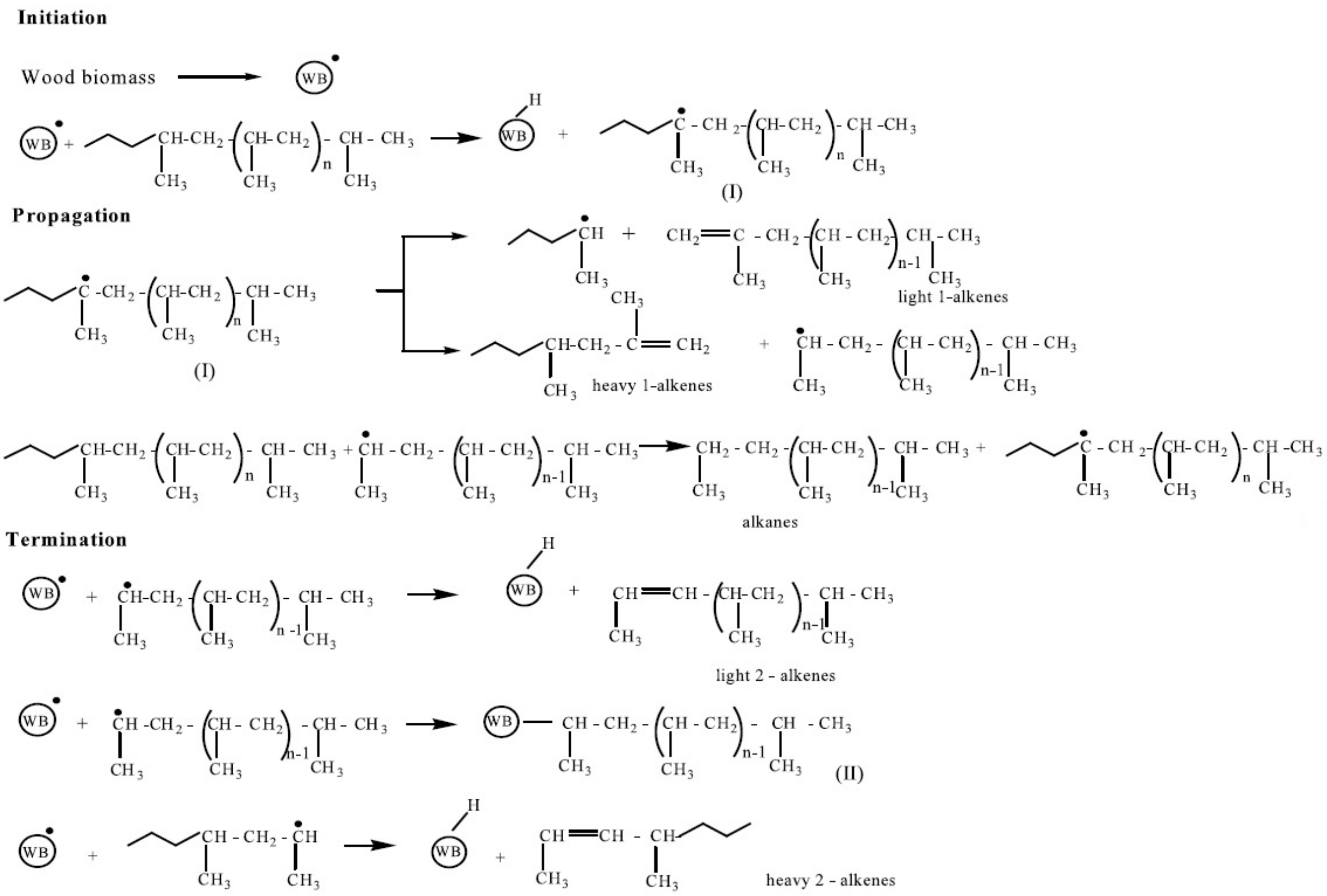

Figure 8. Proposed reaction mechanism for biomass-polymer co-pyrolysis process. Reprinted from (Sharypov et al., 2003), with permission from Elsevier.

### 3.3. Modeling

The best ML model was selected based on four statistical parameters obtained in training and testing. The statistical parameters were correlation coefficient ($R^2$), mean absolute error (MAE), and root mean square error (RMSE). Figure 8 depicts the variations in statistical parameters of the trained ML models. It is evident that the GPR model outperformed other ML models in terms of the statistical parameters considered. The GPR provided the best $R^2$, MAE, and RMSE values in the training and testing phases. These findings could be because of the excellent capacity of the GPR to deal with noisy data and nonlinear systems (Jiang et al., 2021;

Shafizadeh et al., 2022). The average $R^2$ scores of the GPR approach for pyrolysis oil, char, and syngas in the training phase were 0.960, 0.972, and 0.905, respectively. These values were 0.975, 0.986, and 0.916 during the testing phase. The MAE and RMSE values of the GPR model during the training phase varied from $1.7\times10^{-2}$ to $3.14\times10^{-2}$ and from $3.16\times10^{-2}$ to $6\times10^{-2}$ respectively. These values were in the range of $1.56\times10^{-2}$–$2.97\times10^{-2}$ and $2.29\times10^{-2}$–$4.88\times10^{-2}$ during the testing phase, respectively. Generally speaking, higher $R^2$ values (close to unity) and lower error values (MAE and RMSE) introduced the GPR model as a strong model in this study. The hyperparameters of the GPR model optimized by the PSO algorithm are provided in Table 2.

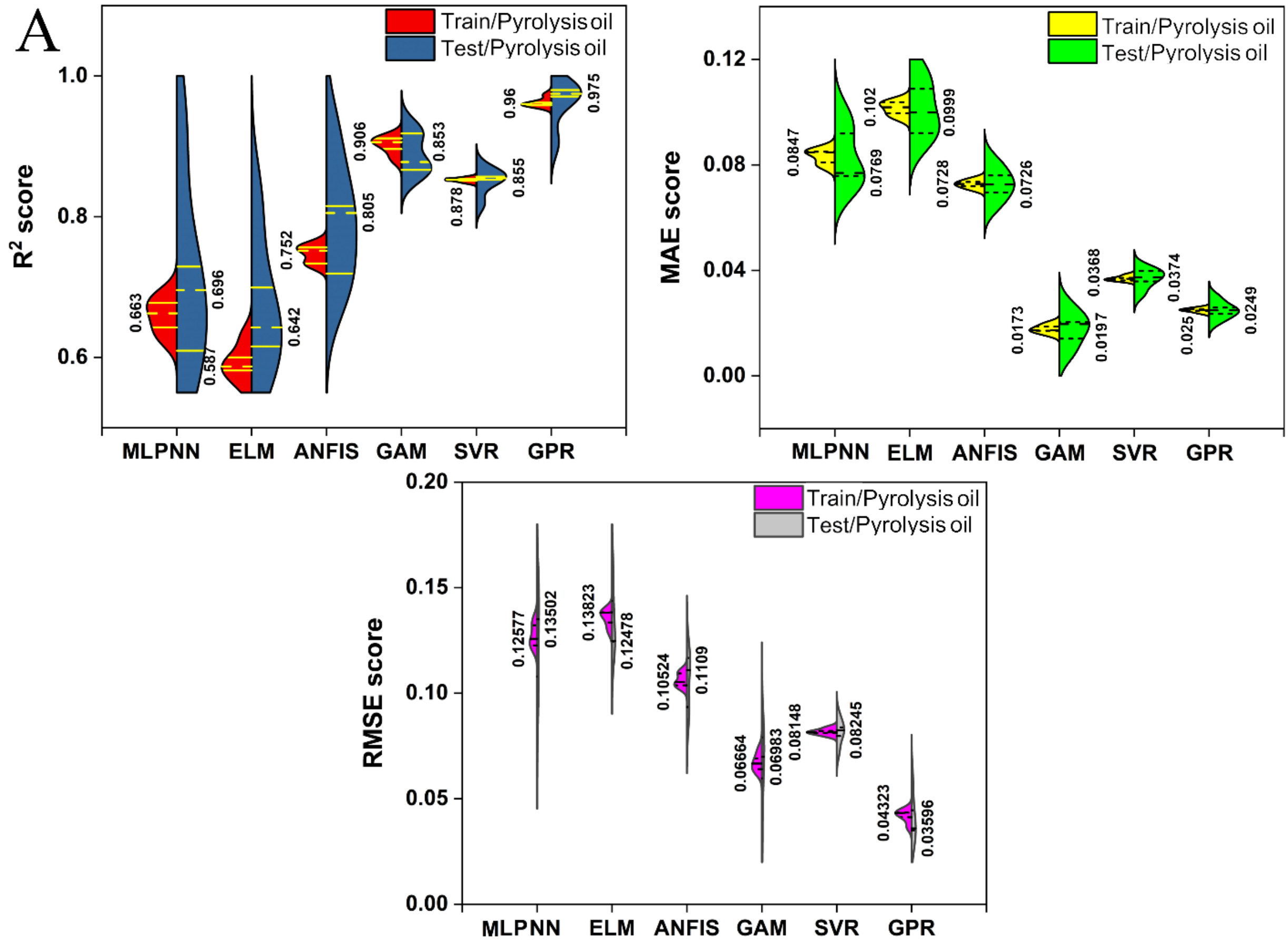

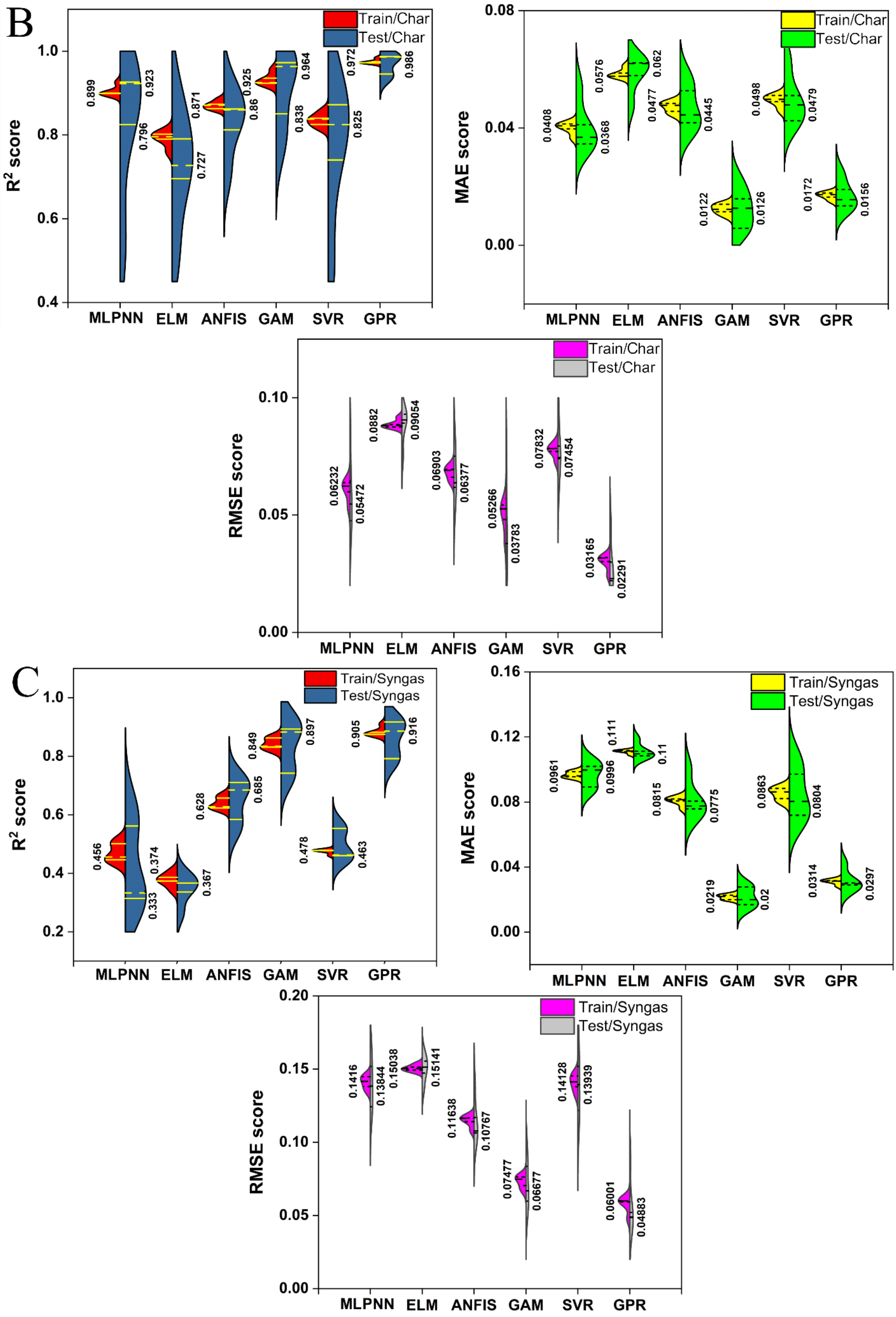

Figure 8. Statistical analysis of the trained ML models for (A) pyrolysis oil yield, (B) char yield, and (C) syngas yield.

Table 2. Hyperparameters of the GPR model optimized by the PSO algorithm.

| Hyperparameter | Pyrolysis oil yield | Char yield | Syngas yield |
|---|---|---|---|
| Kernel Type | ARD Rational Quadratic | ARD Rational Quadratic | ARD Rational Quadratic |
| Scale-mixture parameter | 1.174 | 1.158 | 1.158 |
| Signal standard deviation (SigmaF) | $1.60\times10^{-1}$ | $1.39\times10^{-1}$ | $1.39\times10^{-1}$ |
| Beta | 0.399 | 0.375 | 0.372 |
| Basis functions | Linear | Linear | Linear |

Figure 9 shows the distribution of model-predicted data versus actual data for pyrolysis oil, char, and syngas in the training and testing phases for the 3-fold cross-validation. A similar trend was achieved in the other cross-validation folds (1, 2, 4, and 5). The blue and red lines for the training and testing phases show the regression prediction lines with a 95% confidence interval. The accuracy of the GPR model in predicting char yield was better than pyrolysis oil and syngas yields. However, the accuracy of pyrolysis oil prediction was very close to char. However, the accuracy of the GPR model to predict char yield was much better than syngas yield. The distribution of data points was around the median value for all the outputs using the GPR model (Figure 9F). The GPR-predicted values are well-fitted with the actual data. A linear association between the actual data and the GPR prediction values could be observed. As widely used regression models, MLPNN, ELM, and ANFIS models could not predict the outputs with acceptable accuracy. This issue could be related to the complexity, nonlinearity, and dimensionality of the biomass-polymer co-pyrolysis process. The GAM and SVR models could perform better than MLPNN, ELM, and ANFIS techniques. In general, the GPR model developed could precisely prognosticate the product distribution of the biomass-polymer co-pyrolysis process. Table 3 compares the performance of the ML models developed in this study versus previously published literature. GPR exhibited similar prediction power to previously published models. Moreover, the model developed herein could predict product distribution in biomass-polymer co-pyrolysis.

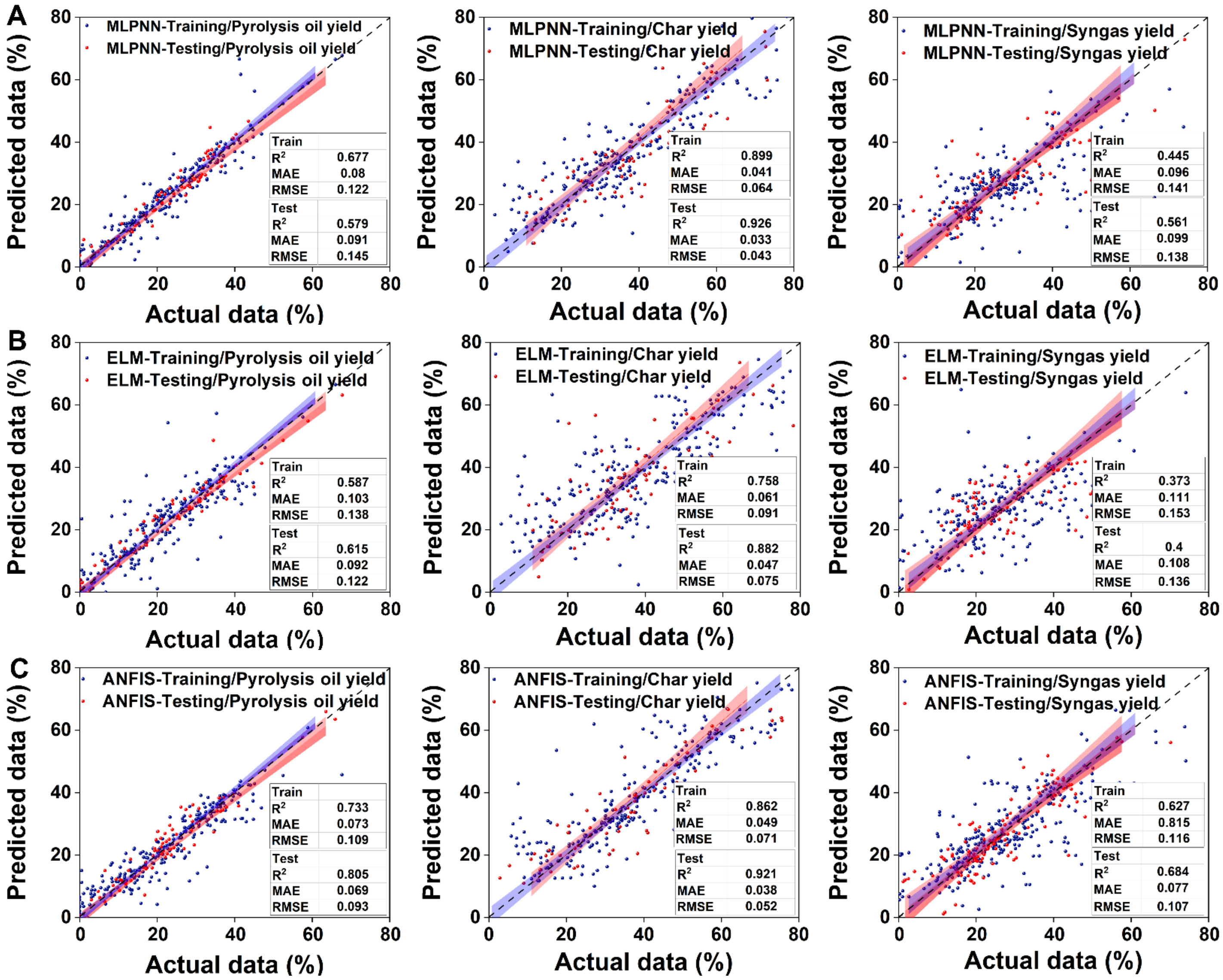
A
B
C
MLPNN-Training/Pyrolysis oil yield
MLPNN-Testing/Pyrolysis oil yield
Train R2 0.677 MAE 0.08 RMSE 0.122
Test R2 0.579 MAE 0.091 RMSE 0.145
MLPNN-Training/Char yield
MLPNN-Testing/Char yield
Train R2 0.899 MAE 0.041 RMSE 0.064
Test R2 0.926 MAE 0.033 RMSE 0.043
MLPNN-Training/Syngas yield
MLPNN-Testing/Syngas yield
Train R2 0.445 MAE 0.096 RMSE 0.141
Test R2 0.561 MAE 0.099 RMSE 0.138
ELM-Training/Pyrolysis oil yield
ELM-Testing/Pyrolysis oil yield
Train R2 0.587 MAE 0.103 RMSE 0.138
Test R2 0.615 MAE 0.092 RMSE 0.122
ELM-Training/Char yield
ELM-Testing/Char yield
Train R2 0.758 MAE 0.061 RMSE 0.091
Test R2 0.882 MAE 0.047 RMSE 0.075
ELM-Training/Syngas yield
ELM-Testing/Syngas yield
Train R2 0.373 MAE 0.111 RMSE 0.153
Test R2 0.4 MAE 0.108 RMSE 0.136
ANFIS-Training/Pyrolysis oil yield
ANFIS-Testing/Pyrolysis oil yield
Train R2 0.733 MAE 0.073 RMSE 0.109
Test R2 0.805 MAE 0.069 RMSE 0.093
ANFIS-Training/Char yield
ANFIS-Testing/Char yield
Train R2 0.862 MAE 0.049 RMSE 0.071
Test R2 0.921 MAE 0.038 RMSE 0.052
ANFIS-Training/Syngas yield
ANFIS-Testing/Syngas yield
Train R2 0.627 MAE 0.815 RMSE 0.116
Test R2 0.684 MAE 0.077 RMSE 0.107
Predicted data (%)
Actual data (%)
0
20
40
60
80

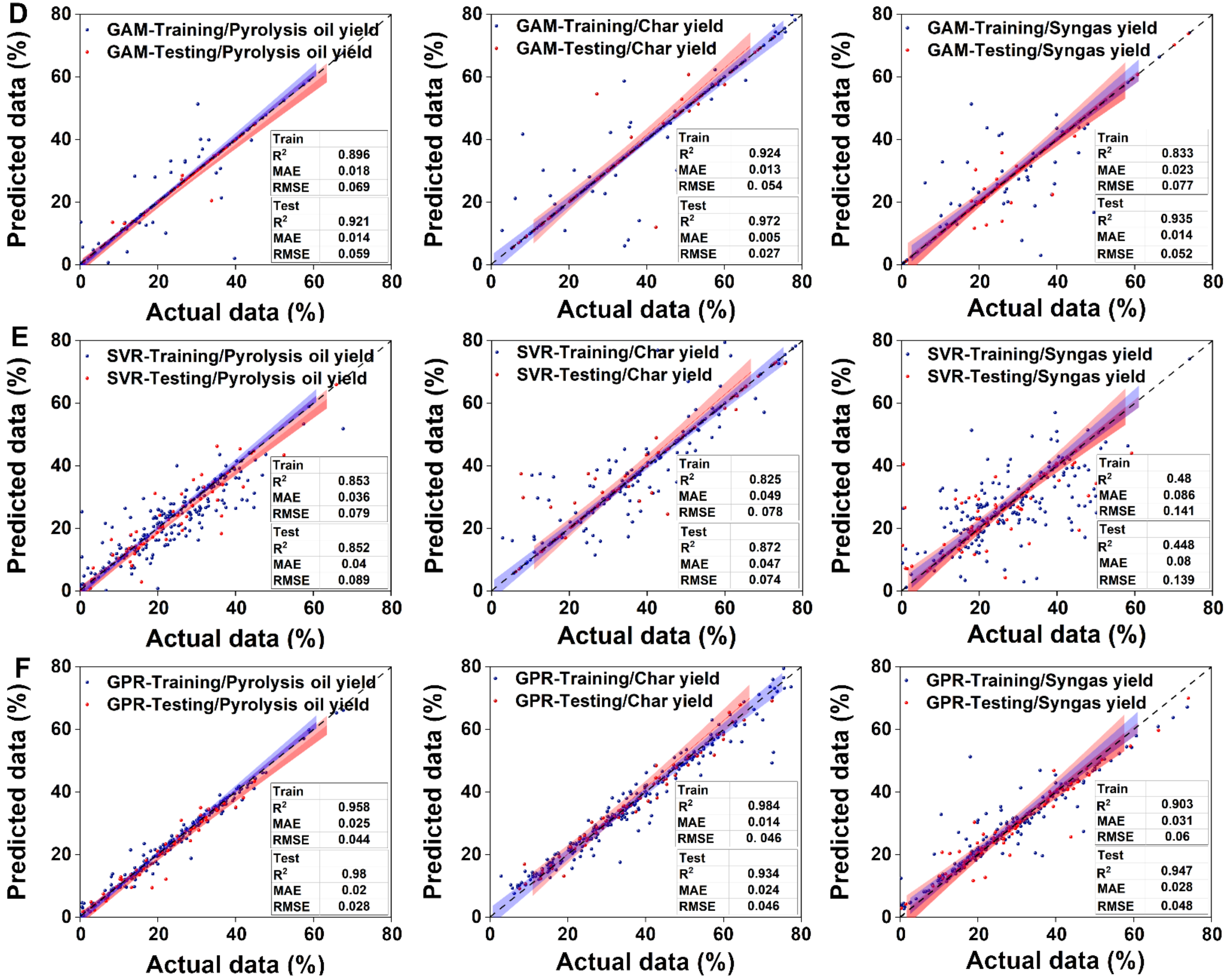

Figure 9. Comparing model-derived values with actual data in the 3-fold cross-validation. (A) MLPNN, (B) ELM, (C) ANFIS, (D) GAM, (E) SVR, and (F) GPR.

Table 3. Comparing the prediction power of the developed models in the current study with previously published literature.

| Ref. | Feedstock type(s) | Best ML model(s) | Model inputs | Model Outputs | Statistical parameter(s) |
| --- | --- | --- | --- | --- | --- |
| Prasertpong et al. (2023) | Biomass feedstocks and plastic wastes | Extreme gradient boosting | Temperature, reaction time, gas conditioning flow rate, heating rate, blending ratio, sample mass loading, sample size, plastic and biomass ultimate analysis (CHNSO), biomass proximate analysis (ash, fixed carbon, volatile matter), | Pyrolytic oil yield, synergic energy | $0.8 < R^2 < 0.88$<br>$0.11 < RMSE < 17.46$ |
| Yang et al. (2023) | Bamboo sawdust and low-density polyethylene | Long short-term memory | Not mentioned | Co-pyrolysis kinetics | $0.9 < R^2 < 0.99$<br>$0.016 < MSE < 0.024$ |
| Alabdrabalnabi et al. (2022) | Biomass feedstocks and plastics | Extreme gradient boosting, neural network | Temperature, reaction time, heating rate, blending ratio, plastic and biomass ultimate analysis (CHNSO), biomass proximate analysis (ash, fixed carbon, volatile matter), | Pyrolytic oil yield, char yield | $R^2 \geq 0.94$<br>$1.77 \leq RMSE \leq 3.26$<br>$1.34 \leq MAE \leq 2.6$ |
| Present study | Biomass feedstocks and polymeric wastes | GPR | Temperature, reaction time, heating rate, blending ratio, both plastic and biomass ultimate analysis (CHNSO), biomass proximate analysis (ash, fixed carbon, volatile matter), | Pyrolytic oil yield, char yield, syngas yield | $0.93 < R^2 < 0.98$<br>$0.0316 \leq RMSE \leq 0.06$<br>$0.017 \leq MAE \leq 0.034$ |

### 3.4. Process optimization

The MOPSO algorithm was employed to find the optimal biomass and polymer waste composition and operating conditions for maximizing pyrolysis oil yield during the co-pyrolysis process. The optimal points and their respective outputs provided by the MOPSO algorithm are presented in the "Optimization results" sheet in the "Supplementary Material". The maximum pyrolysis oil yield was obtained when the carbon, hydrogen, nitrogen, sulfur, and oxygen contents of biomass (polymer) were in the range of 68.3–79.8% (92.2–92.4%), 2.20–4.64% (8.86–13.0%), 1.16–4.53% (0.00–0.74%), 19.1–29.6% (0.00–0.23%), and 0.00–3.87% (0.00–0.3%), respectively. Under the selected optimal conditions, the volatile matter, fixed carbon, and ash content biomass (polymer) were 96.0–97.9% (99.5–100), 1.10–16.9% (0.00–6.23%) and 0.07–1.93% (0.00–0.68%), respectively. The optimum reaction conditions were: biomass blending percentage of 18–27%, a temperature of 509–629 °C, a heating rate of 5–64 °C/min, and a reaction time of 33–118 min. The pyrolysis oil yield reached 70.9–75.3% under optimum conditions. The char and syngas yields were 7.23–21.5% and 5.68–18.6%, respectively, under the selected optimal conditions. The optimization process by MOPSO could greatly maximize the pyrolysis oil yield while minimizing the other two byproducts. The results revealed that biomass and polymer waste with higher carbon content and volatile matter could increase pyrolysis oil yield in the co-pyrolysis process. In addition, feedstocks with more polymeric wastes and less biomass composition (3 to 1 proportion) could improve pyrolysis oil production. The optimal operating conditions (temperature, heating rate, and reaction time) obtained herein could be applied in the co-pyrolysis process to maximize pyrolysis oil yield.

### 3.5. Challenges and future directions

Even though this study demonstrates promising results, more accurate and robust ML models are still needed to predict biomass-polymer co-pyrolysis. By compiling a more comprehensive database, this issue can be accomplished. Furthermore, the developed model is not suitable for catalytic biomass-polymer co-pyrolysis. Therefore, the effects of catalyst parameters (such as metal, support, and promoter types and their loading) should be included in future studies. A lack of sufficient reported data in the literature prevented the presented model from covering the quality of pyrolytic oil. The most important qualitative parameters are the chemical composition (the quantity of acids, esters, and phenols) and physical properties (viscosity, density, and heating value) of pyrolytic oil. If enough experimental data could be gathered from the literature, this limitation could be overcome in future studies. Based on the results, the selected model is a viable alternative to labor-intensive and expensive co-pyrolysis experiments. This model could also be used in the technical-economic-environmental analysis of the biomass-polymer co-pyrolysis process. In addition, biomass-polymer co-pyrolysis could be accurately monitored and controlled using such a precise ML model. The optimal conditions obtained herein could be used in real-world situations to maximize pyrolysis oil yield.

## 4. Conclusions

This study introduced an evolutionary ML model to predict the product distribution of the biomass-polymer co-pyrolysis process. An inclusive database covering an extensive range of biomass and polymer compositions under various operating conditions was prepared from the published literature. The extracted data were reconstructed using an innovative approach to effectively reflect the nature of biomass and polymer on the output parameters. The PCA technique was used to decrease the dimensionality of the dataset and select the applicable information. The PCA outputs were normalized and introduced into six ML models (i.e., MLPNN, ELM, ANFIS, GAM, SVR, and GPR). The GPR model developed revealed an outstanding prediction capability with an $R^2$ value higher than 0.90. In addition, the statistical errors (MAE and RMSE) of the GPR model for all the outputs were remarkably low. The biomass-polymer co-pyrolysis process was optimized using the MOPSO method by the GPR-derived objective functions. The goal was to maximize pyrolysis oil yield while minimizing syngas and char yields. Under optimal conditions, the pyrolysis oil yield was as high as 70.9–75.3%. The char and syngas yields were 7.23–21.5% and 5.68–18.6%, respectively.